\documentclass[letterpaper, 10 pt, conference]{IEEEtran}

\usepackage[pdftex]{graphicx}

\usepackage{amsmath}
\usepackage{amssymb}
\usepackage{mathtools}

\newcommand{\crossmat}[1]{\left[#1\times\right]}
\newcommand{\DotHat}[1]{\Dot{\hat{#1}}}
\newcommand{\eye}[1]{\mathbf{I}_{#1}}
\newcommand{\zeros}[2]{\mathbf{0}_{#1 \times #2}}
\newcommand{\quatRot}[1]{\mathcal{C}(#1)}
\newcommand{\quatJac}[2]{\mathcal{J}(#1,#2)}

\usepackage{cite}
\usepackage{hyperref}
\usepackage[capitalise, noabbrev]{cleveref}

\usepackage{algorithmicx}

\usepackage{array}
\usepackage{flushend}

\IEEEoverridecommandlockouts
\usepackage{tikz}
\usepackage{textcomp}
\usepackage{lipsum}

\newcommand\copyrighttext{%
  \footnotesize \textcopyright 2022 IEEE. Personal use of this material is permitted.
  Permission from IEEE must be obtained for all other uses, in any current or future
  media, including reprinting/republishing this material for advertising or promotional
  purposes, creating new collective works, for resale or redistribution to servers or
  lists, or reuse of any copyrighted component of this work in other works.
  DOI: \href{https://doi.org/10.1109/SSRR56537.2022.10018692}{10.1109/SSRR56537.2022.10018692}}
\newcommand\copyrightnotice{%
\begin{tikzpicture}[remember picture,overlay]
\node[anchor=south,yshift=10pt] at (current page.south) {\fbox{\parbox{\dimexpr\textwidth-\fboxsep-\fboxrule\relax}{\copyrighttext}}};
\end{tikzpicture}%
}

\begin{document}

\title{Online Multi Camera-IMU Calibration}

\author{Jacob~Hartzer and Srikanth~Saripalli}

\maketitle
\copyrightnotice

\begin{abstract}
    Visual-inertial navigation systems are powerful in their ability
    to accurately estimate localization of mobile systems within complex
    environments that preclude the use of global navigation satellite systems.
    However, these navigation systems are reliant on accurate and
    up-to-date temporospatial calibrations of the sensors being used. As such,
    online estimators for these parameters are useful in resilient
    systems. In this paper, we present an extension to existing Kalman Filter
    based frameworks for estimating and calibrating the extrinsic parameters of
    IMU and multi-camera systems. In addition to extended the filter framework
    to include multiple camera sensors, we reformulate the measurement model to
    make use of measurement data that is typically available in fiducial
    detection software. Finally, we include a secondary set of filters that
    estimate the time translation parameters without closed-loop feedback.
    Experimental calibration results, including cameras with non-overlapping
    fields of view. Finally the code has been open-sourced and made available.
\end{abstract}


%
\IEEEpeerreviewmaketitle

\section{Introduction}
%
%
%
%

For any mobile robotic system, localization is key to the success of a
mission. The ability for a vehicle or robot to precisely determine its location
within a local reference frame is paramount to the subsequent problems of
navigation, path planning, and control. Historically, high accuracy localization
was achieved using a combination of an Inertial Measurement Unit (IMU) and
a Global Navigation Satellite System (GNSS), colloquially referred to
as GPS. The success of this combination of sensors came from the high update
rates achievable by IMUs with relatively good accuracy over short time periods,
and the zero drift corrections coming from the GPS.

Unfortunately, GPS aided navigation is often unavailable or inaccurate due to
environmental factors. Any system that is indoors, underwater, or surrounded by
tall buildings would be unable to reliably use GPS for localization.
Additionally, high accuracy GPS receivers are often too large and expensive to
deploy in large numbers for small robotics. Comparatively, cameras are
lightweight, inexpensive, and provide rich environmental information. As such,
Visual Inertial Navigation Systems (V-INS) have sought to leverage the robust
information and combine these measurements with inertial sensors to provide high
speed and high accuracy localization \cite{OpenVINS,xVIO}.

\begin{figure}[h]
    \centering
    \includegraphics[width = 0.9\linewidth]{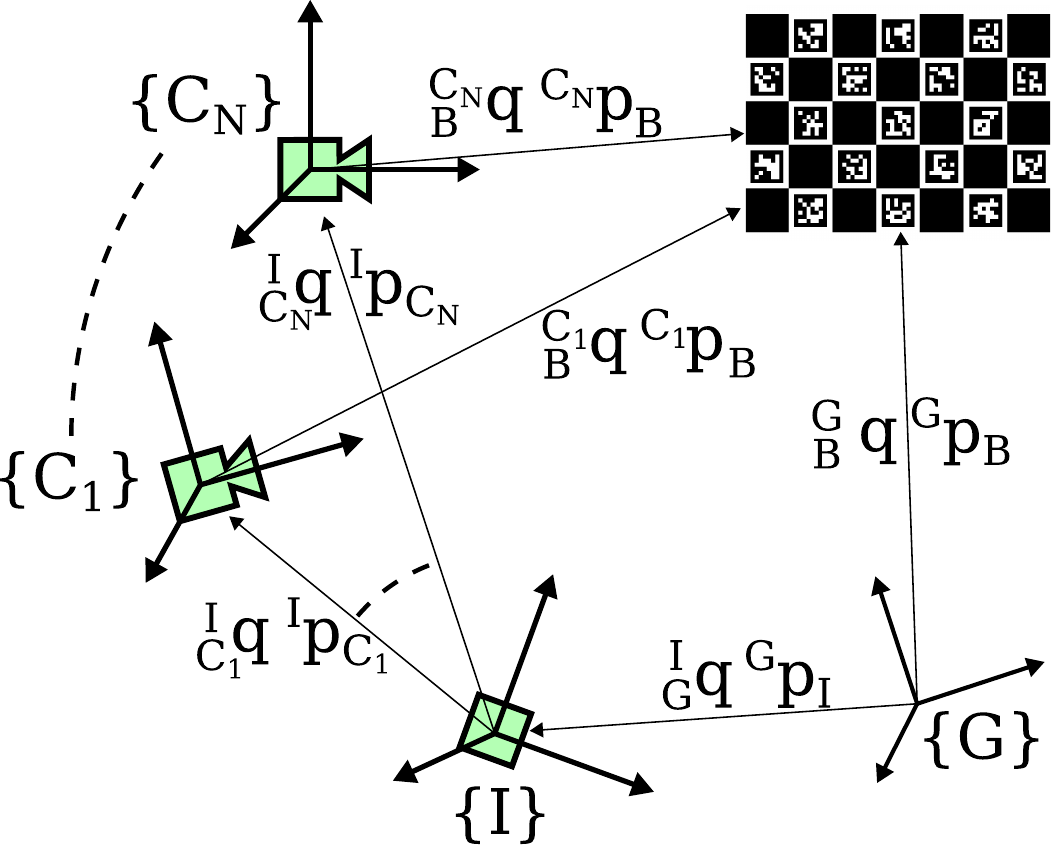}
    \caption{
        Relationship graph between frames of reference showing
        translations and rotations between the Global \{$G$\}, Board \{$B$\},
        IMU \{$I$\}, and Camera \{$C_1$...$C_N$\} frames
    }
    \label{fig:setup}
\end{figure}

For a V-INS to work properly, it is necessary to accurately calibrate camera IMU
systems for high-accuracy localization. Methods for performing this calibration
include the use of high accuracy 3D measurements using CAD models or 3D scanners.
However, these methods can be prohibitive due to cost and resource constraints.
Additionally, without further calibration of the physical sensor body with the
produced measurements, these types of calibrations cannot fully calculate the
extrinsic calibration parameters. Finally, due to shifts in sensor mounting,
maintenance, or repair, calibrations cannot be assumed to be constant over the
lifetime of a mobile system, further increasing the costs of calibration.

Because of these limitations, offline batch optimization has often been used to
calibrate systems using measurement data in a defined environment. This method
can produce accurate extrinsic and temporal calibration parameters of camera-IMU
systems with very little cost
\cite{Furgale2013,Tan2013,Fleps2011}.
However, this calibration must be
done as a non-sequential batch process which would struggle with estimating
parameters over a large time scale. Additionally, batch processes by their
nature are unable to handle sudden changes to sensor extrinsics, such as a
sensor shifts due to disturbance or vibration.

As such, online calibrations using Extended Kalman Filter (EKF) frameworks have
been used to calibrate camera-IMU systems
\cite{Mirzaei2007,Hu2020,Eckenhoff2021,Parnian2010,Guodong2009a}. These Kalman
filter based methods can perform online calibrations as well as handle step
changes to sensor extrinsic parameters. In this paper, we present an update to
the EKF-based algorithm for IMU-camera system calibration that incorporates
multiple cameras. This framework adapts existing measurement models to take
advantage of open source computer vision software for the detection of visual
fiducials, which can lead to decreased developmental complexity and simplify the
filter measurement update. We also provide an example of this framework through
the open-sourced repository found at \cite{multi-cam-imu-cal}. Finally, we
consider temporal calibration through the development of a simple one way time
translation filter that can be applied to any sensor measurement that does not
utilize two-way clock synchronization.

\section{Related Work}

The IMU and single camera calibration filter that this paper extends and adapts
was originally proposed and implemented in \cite{Mirzaei2007}.
Reformulations of this filter have been successfully implemented
using unscented Kalman filters \cite{Brink, Kelly} or a multi-state constraint
Kalman filter \cite{Hu2020}. We extend the results
of what has been previously performed through the experimental use of multiple
cameras with differing data transmission types and rates, as well as the use
of quaternion measurements in the camera update function. This is done because
ArUco markers can be assumed to be a 3D pose sensor with low error in both
relative position and orientation \cite{Garrido-Jurado2014}. This sensing model
is used in place of the 2D projective model previously used in other online
filter frameworks \cite{Mirzaei2007}. This was done to show how any fiducial
marker that satisfies the simplifying assumption of being a robust 3D sensor
could be used to quickly incorporate a camera into a localization filter.

Calibration checkerboards are often used for their subpixel accuracy when
calibrating camera intrinsics. Combining checkerboards with fiducials such as
ArUco markers these two in the form of ChArUco markers allows for subpixel
accuracy as well as the discerning of 180 degree rotations. The added simplicity
of using open source fiducial marker detection packages such as OpenCV allows
for rapid development \cite{opencv}. As such, we also show that it is possible
to make use of more complex fiducials that can offer more accurate measurements
than simple feature tracking. By taking any existing fiducial detector, the
framework described in this work is capable of quickly adopting any number
styles of marker and associated detectors.

The calibration of proprioceptive and exteroceptive sensors is often considered
a spatial problem, where intrinsic properties are calculated separately from the
extrinsic, and the main concern is in determining a translation vector and
rotation matrix. This manifests in work where the goal is to determine these
extrinsic parameters either through online or offline methods
\cite{Park2012, Kroeger2019}.
However, the calibration problem is not purely spatial, but rather
\textit{temporospatial} inasmuch as it is necessary to know the location,
orientation and time which a measurement was taken in order to properly include
in a filter in an optimal way. As such, work has been done
to more rigorously calibrate sensors temporally using both offline
and online methods \cite{Kelly}. Additionally, the relative importance of
accurate time measurements has been shown to have significant impacts on
exteroceptive sensors such as cameras \cite{TriggerSync}. While others have
solved this temporospatial calibration by including the time translation into
the Kalman state vector, we propose a sub-filter methodology, that is not
reliant on state updates to correct sensor timings. This can be applied to any
sensor that cannot take advantage of closed loop synchronization systems such as
Precision Time Protocol (PTP) or where hardware time synchronization is either
difficult or impossible to achieve with given sensor
hardware \cite{Sommer2017}.

\section{Algorithm Description}

The filter state is composed of a singular IMU state and multiple independent
camera states represented by $\boldsymbol{x}_{imu}$ and $\boldsymbol{x}_{cam_i}$
respectively. The IMU state contains
the rotation of the global frame in the IMU frame $\prescript{I}{G}{q}$,
the gyroscope bias $\mathbf{}{\mathbf{b}}_g$,
the imu velocity in the global frame $\prescript{G}{}{\mathbf{v}}_I$,
the accelerometer bias $\mathbf{b}_a$,
and the imu position in the global frame $\prescript{G}{}{\mathbf{p}}_I$.
Additionally, each camera state is composed of
the orientation in the IMU frame $\prescript{I}{C}{q}$,
and the position in the IMU frame $\prescript{I}{}{\mathbf{p}}_C$.
Shown in \cref{fig:setup} are the relative frames for the Global \{$G$\},
Calibration Board \{$B$\}, IMU \{$I$\}, and Cameras \{$C_1$\}...\{$C_N$\}.
Additionally shown are the inter-frame translations and rotations $p$ and $q$,
respectively. Therefore, the entire state matrix with $N$ cameras is

\begin{equation}
    \boldsymbol{x} =
    \begin{bmatrix}
        \boldsymbol{x}_{imu}   &
        \boldsymbol{x}_{cam_1} & ... &
        \boldsymbol{x}_{cam_N}
    \end{bmatrix}^T
\end{equation}
where
\begin{equation}
    \boldsymbol{x}_{imu} =
    \begin{bmatrix}
        \prescript{I}{G}{q}           &
        \mathbf{}{\mathbf{b}}_g       &
        \prescript{G}{}{\mathbf{v}}_I &
        \mathbf{b}_a                  &
        \prescript{G}{}{\mathbf{p}}_I
    \end{bmatrix}
\end{equation}
\begin{equation}
    \boldsymbol{x}_{cam_i} =
    \begin{bmatrix}
        \prescript{I}{C}{q} &
        \prescript{I}{}{\mathbf{p}}_C
    \end{bmatrix}_i
\end{equation}

\subsection{One-Way Time Translation}

A one-way time translation filter was created using a first order translation
model, similar to the formation in \cite{TriggerSync}. We assume the map from
the measured sensor time $t_s$ to the computer time $t_c$ is linear with a skew
$\alpha$ and offset $\beta$.
\begin{equation} \label{eq:time_trans}
    t_c = \alpha \cdot t_s + \beta
\end{equation}
With this model, the filter state is composed of the translation equation coefficients
\begin{equation}
    \mathbf{x} =
    \begin{bmatrix}
        \alpha & \beta
    \end{bmatrix}^T
\end{equation}
We initialize the filter using the initial sensor time $t_{c_0}$ and initial
main computer time $t_{c_0}$.
\begin{equation}
    \mathbf{x}_0 =
    \begin{bmatrix}
        1 \\ t_{c_0} - t_{s_0}
    \end{bmatrix}
\end{equation}
The measurement $z$ and measurement model $\hat{z}$ are simply
\begin{equation}
    z = t_{c_0} - t_{s_0}
\end{equation}
\begin{equation}
    \hat{z} = \alpha \cdot t_s + \beta
\end{equation}
Therefore, the observation matrix is
\begin{equation}
    \mathbf{H} =
    \begin{bmatrix}
        t_s & 1
    \end{bmatrix}
\end{equation}
With these definitions, the typical EKF equations are used to
update the filter state and covariance. Using the filter state values with
\cref{eq:time_trans} gives an estimate of the measurement time translated to
the main computer frame of reference.

\begin{figure}
    \centering
    \includegraphics[width = 1.0\linewidth]{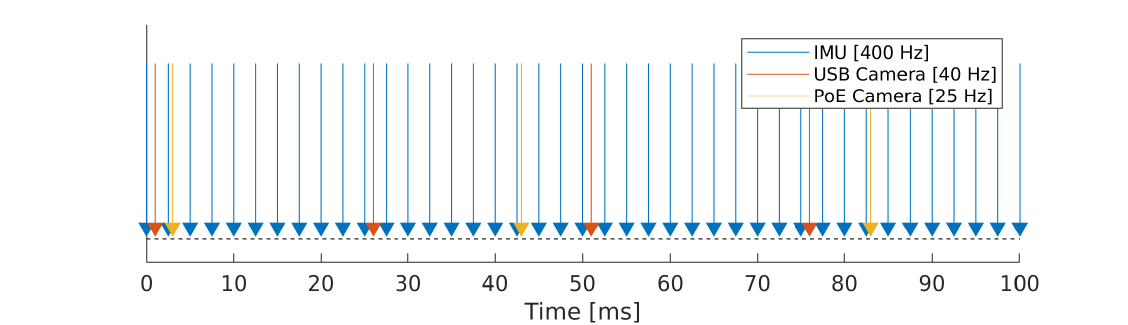}
    \caption{Sensor update rates used in the experimental work}
    \label{fig:rateDiagram}
\end{figure}

These filters were created outside the main calibration filter in this work due
to a number of reasons. First, because the highly uncorrelated nature of the
timing measurements from different sensors, it would unnecessarily increase the
complexity of the filter to include each $\alpha$ and $\beta$ parameter for
every sensor using time translation. Additionally, this method allows for
greater flexibility of the applied time-translation filter. In the case of this
work, one camera utilized PTP while the other utilized a time-translation
filter with update rates outlined in \cref{fig:rateDiagram}. As such, the
separation of the filters allows the extrinsic camera parameters to be
vectorized in the main filter. Finally, because the time a sensor measurement is
taken is relevant to the update it provides, this method provides updated sensor
measurement time before it is used within the main calibration filter.

\subsection{Calibration State Initialization}

The filter state is initialized using initial estimates of the extrinsic
parameters and the first update. The initial extrinsic parameter estimates
can come from any calibration source. The IMU pose is initialized
upon the first successful camera measurement of
$(\prescript{B}{}{\mathbf{p}}_C, ~\prescript{B}{C}{q})$ as follows
\begin{equation}
    \begin{split}
        \prescript{I}{G}{\hat{q}} &= (
        \prescript{G}{B}{q} \otimes
        \prescript{B}{C}{q} \otimes
        \prescript{C}{I}{\hat{q}} )^{-1}
        \\
        \prescript{G}{}{\hat{\mathbf{p}}}_I &=
        \prescript{G}{}{\mathbf{p}}_C +
        \prescript{G}{}{\mathbf{p}}_B -
        \quatRot{ \prescript{I}{G}{\hat{q}}}^T
        \prescript{I}{}{\mathbf{p}}_C
    \end{split}
\end{equation}
where $\quatRot{q}$ represents the rotation matrix corresponding to
quaternion $q$ and $\otimes$ represents quaternion multiplication.

\subsection{Calibration State Propagation}

As shown in \cite{Mirzaei2007}, we can apply the expectation operator to the
continuous-time dynamics of the system, and produce the following state
propagation equations. The only notable difference is the expansion of these
propagation equations to include multiple cameras, denoted by the subscript $N$
for each camera reference frame $C$.
\begin{align}
    \prescript{I}{G}{\Dot{\hat{q}}}(t)         & = \frac{1}{2}  \Omega(\hat{\boldsymbol{\omega}}(t))\prescript{I}{G}{\hat{q}}(t) \label{eq:prop1} \\
    \DotHat{\mathbf{b}}_g(t)                   & = \mathbf{0}_{3 \times 1} \label{eq:prop2}                                                       \\
    \prescript{G}{}{\DotHat{\mathbf{v}}}_I(t)  & = \boldsymbol{C}^T( \prescript{I}{G}{ \hat{q} }(t) )
    \prescript{G}{}{\hat{\mathbf{a}}}(t) + \prescript{G}{}{\mathbf{g}} \label{eq:prop3}                                                           \\
    \DotHat{\mathbf{b}}_a(t)                   & = \mathbf{0}_{3 \times 1} \label{eq:prop4}                                                       \\
    \prescript{G}{}{\DotHat{\mathbf{p}}}_I(t)  & = \prescript{G}{}{\hat{\mathbf{v}}}_I(t) \label{eq:prop5}                                        \\
    \prescript{I}{{C_N}}{\DotHat{q}}(t)        & = \mathbf{0}_{3 \times 1} \label{eq:prop6}                                                       \\
    \prescript{I}{}{\Dot{\mathbf{p}}}_{C_N}(t) & = \mathbf{0}_{3 \times 1} \label{eq:prop7}
\end{align}

By integrating
\cref{eq:prop1,eq:prop2,eq:prop3,eq:prop4,eq:prop5,eq:prop6,eq:prop7}, a
propagated state estimate is produced in discrete time. However, in order to
propagate the state covariance, an error state filter using multiple cameras
is used where the error state is
\begin{equation}
    \boldsymbol{\widetilde{x}} =
    \begin{bmatrix}
        \boldsymbol{\widetilde{x}}_{imu}   &
        \boldsymbol{\widetilde{x}}_{cam_1} & ... &
        \boldsymbol{\widetilde{x}}_{cam_N}
    \end{bmatrix}^T
\end{equation}
where
\begin{equation}
    \boldsymbol{\widetilde{x}}_{imu} =
    \begin{bmatrix}
        \prescript{I}{G}{\delta \theta}               &
        \widetilde{\mathbf{b}}_g                      &
        \prescript{G}{}{ \widetilde{ \mathbf{v} } }_I &
        \widetilde{\mathbf{b}}_a                      &
        \prescript{G}{}{ \widetilde{ \mathbf{p} } }_I
    \end{bmatrix}
\end{equation}
\begin{equation}
    \boldsymbol{\widetilde{x}}_{cam_i} =
    \begin{bmatrix}
        \prescript{I}{C}{\delta \theta} &
        \prescript{G}{}{ \widetilde{ \mathbf{p} } }_C
    \end{bmatrix}_i
\end{equation}
The linearized continuous-time error-state propagation is given by
\begin{equation}
    \Dot{\widetilde{\mathbf{x}}} = \mathbf{F} \widetilde{\mathbf{x}} + \mathbf{G} \mathbf{n}
\end{equation}
where for N cameras
\begin{equation*}
    \mathbf{F} =
    \begin{bmatrix}
        -\crossmat{\Hat{\boldsymbol{\omega}}}                           & -\eye{3}      & \zeros{3}{3}  & \zeros{3}{3}                           & \zeros{3}{9}  \\
        \zeros{3}{3}                                                    & \zeros{3}{3}  & \zeros{3}{3}  & \zeros{3}{3}                           & \zeros{3}{9}  \\
        \quatRot{\prescript{I}{G}{\hat{q}}} \crossmat{\hat{\mathbf{a}}} & \zeros{3}{3}  & \zeros{3}{3}  & -\quatRot{\prescript{I}{G}{\hat{q}}}^T & \zeros{3}{9}  \\
        \zeros{3}{3}                                                    & \zeros{3}{3}  & \zeros{3}{3}  & \zeros{3}{3}                           & \zeros{3}{9}  \\
        \zeros{3}{3}                                                    & \zeros{3}{3}  & \eye{3}       & \zeros{3}{3}                           & \zeros{3}{9}  \\
        \zeros{6N}{3}                                                   & \zeros{6N}{3} & \zeros{6N}{3} & \zeros{6N}{3}                          & \zeros{6N}{9} \\
    \end{bmatrix}
\end{equation*}
\begin{equation*}
    \mathbf{G} =
    \begin{bmatrix}
        -\eye{3}      & \zeros{3}{3}  & \zeros{3}{3}                         & \zeros{3}{3}  \\
        \zeros{3}{3}  & \eye{3}       & \zeros{3}{3}                         & \zeros{3}{3}  \\
        \zeros{3}{3}  & \zeros{3}{3}  & -\quatRot{\prescript{I}{G}{\hat{q}}} & \zeros{3}{3}  \\
        \zeros{3}{3}  & \zeros{3}{3}  & \zeros{3}{3}                         & \eye{3}       \\
        \zeros{3}{3}  & \eye{3}       & \zeros{3}{3}                         & \zeros{3}{3}  \\
        \zeros{6N}{3} & \zeros{6N}{3} & \zeros{6N}{3}                        & \zeros{6N}{3} \\
    \end{bmatrix},
    ~
    \mathbf{n} =
    \begin{bmatrix}
        \mathbf{n}_{g}  \\
        \mathbf{n}_{wg} \\
        \mathbf{n}_{a}  \\
        \mathbf{n}_{wa}
    \end{bmatrix}
\end{equation*}
and the cross product operator is defined as
\begin{equation}
    \crossmat{
        \boldsymbol{\omega}
    } \equiv
    \begin{bmatrix}
        0         & -\omega_z & -\omega_y \\
        \omega_z  & 0         & -\omega_x \\
        -\omega_y & \omega_x  & 0
    \end{bmatrix}
\end{equation}
The remaining derivations are consistent with the typical development of an
error state Kalman filter which is well described in \cite{Sola}.

\subsection{Measurement Model}
\label{sec:MeasurementModel}

The cameras, which are rigidly attached to the IMU, record images of a
calibration target where the target frame has a known location and orientation
within the global frame
$\prescript{G}{}{\mathbf{p}}_B $ and $\prescript{B}{G}{q} $
respectively. Using a calibration target such as a grid board of April tags or
a ChArUco board, it is possible to use the \textit{estimatePoseBoard()} function from OpenCV to
extract translation and rotation vectors in the target frame \cite{opencv}.
Therefore, the measurement model for each camera is simply
\begin{equation}
    \mathbf{z} =
    \begin{bmatrix}
        \prescript{B}{}{\mathbf{p}}_C \\
        \prescript{B}{C}{q}
    \end{bmatrix}
\end{equation}
\begin{equation}
    \hat{\mathbf{z}} =
    \begin{bmatrix}
        \quatRot{\prescript{B}{G}{q}}(
        \prescript{G}{}{\hat{\mathbf{p}}}_I -
        \prescript{G}{}{\mathbf{p}}_B +
        \quatRot{\prescript{B}{G}{\hat{q}}}
        \prescript{I}{}{\hat{\mathbf{p}}}_C ) \\
        \prescript{I}{C}{\hat{q}}^{-1} \otimes
        \prescript{I}{G}{\hat{q}}      \otimes
        \prescript{B}{G}{q}^{-1}
    \end{bmatrix}
\end{equation}
The measurement matrix for the first camera is
\begin{equation} \label{eq:H_1}
    \mathbf{H} =
    \begin{bmatrix}
        \mathbf{\prescript{p}{}{J}}_{\prescript{I}{G}{q},p} &
        \zeros{3}{9}                                        &
        \mathbf{J}_{\prescript{G}{}{p}_I}                   &
        \mathbf{\prescript{p}{}{J}}_{\prescript{I}{C}{q}}   &
        \mathbf{J}_{\prescript{I}{C}{p}}                    &
        \zeros{3}{6}                                          \\
        \mathbf{\prescript{q}{}{J}}_{\prescript{I}{G}{q},q} &
        \zeros{3}{9}                                        &
        \zeros{3}{3}                                        &
        \mathbf{\prescript{q}{}{J}}_{\prescript{i}{C}{q}}   &
        \zeros{3}{3}                                        &
        \zeros{3}{6}
    \end{bmatrix}
\end{equation}
and for the second camera is
\begin{equation} \label{eq:H_2}
    \mathbf{H} =
    \begin{bmatrix}
        \mathbf{\prescript{p}{}{J}}_{\prescript{I}{G}{q},p} &
        \zeros{3}{9}                                        &
        \mathbf{J}_{\prescript{G}{}{p}_I}                   &
        \zeros{3}{6}                                        &
        \mathbf{\prescript{p}{}{J}}_{\prescript{I}{C}{q}}   &
        \mathbf{J}_{\prescript{I}{C}{p}}                      \\
        \mathbf{\prescript{q}{}{J}}_{\prescript{I}{G}{q},q} &
        \zeros{3}{9}                                        &
        \zeros{3}{3}                                        &
        \zeros{3}{6}                                        &
        \mathbf{\prescript{q}{}{J}}_{\prescript{i}{C}{q}}   &
        \zeros{3}{3}
    \end{bmatrix}tem
\end{equation}
where
\begin{align}
    \mathbf{\prescript{p}{}{J}}_{\prescript{I}{G}{q}} & = -\quatRot{\prescript{I}{C}{q}}^T \quatRot{\prescript{I}{G}{q}} \crossmat{\prescript{G}{}{p}_B - \prescript{G}{}{p}_I} \quatJac{\prescript{I}{G}{q}}{} \\
    \mathbf{J}_{\prescript{G}{}{p}_I}                 & = -\quatRot{\prescript{I}{C}{q}}^T \quatRot{\prescript{I}{G}{q}}                                                                                        \\
    \mathbf{\prescript{p}{}{J}}_{\prescript{I}{C}{q}} & = -\quatRot{\prescript{I}{C}{q}}^T \crossmat{  \quatRot{\prescript{I}{G}{q}} * (\prescript{G}{}{p}_B - \prescript{G}{}{p}_I) - \prescript{I}{}{p}_C}    \\
    \mathbf{J}_{\prescript{I}{C}{p}}                  & = -\quatRot{\prescript{I}{C}{q}}^T
\end{align}
Note that the Jacobians
$\mathbf{J}_{\prescript{I}{G}{q},p}$ and $\mathbf{J}_{\prescript{I}{C}{q},q}$
are not reproduced here due to the complexity of the analytical solution. In
practice, the use of numerical solutions for these jacobians is simpler and
computationally faster. Additionally, these Jacobians are not taken with respect
to the measurement $\hat{z}$ due to the singularity at the 180 degree rotation
mark. Rather, we take the Jacobian of the residual $\hat{z} - z$ which produces
much smaller measurements, as is expected with an error state Kalman filter
formulation. These smaller measurements produce more linear and stable results
when the system experiences large rotations.

Moreover, the structure of this
filter allows for N cameras by adjusting the column location of the
measurement matrix as shown in \cref{eq:H_1,eq:H_2} and is therefore quite
extensible in both analysis and in software.

\subsection{Calibration Measurement Update}

Using the measurement model outlined in \cref{sec:MeasurementModel}, we apply
the usual Kalman update equations
\begin{equation}
    \mathbf{y} = \mathbf{z} - \hat{\mathbf{z}}
\end{equation}
\begin{equation}
    \mathbf{S} = \mathbf{H} \mathbf{P} \mathbf{H}^T + \mathbf{R}
\end{equation}
\begin{equation}
    \mathbf{K} = \mathbf{P} \mathbf{H}^T \mathbf{S}^{-1}
\end{equation}
The Mahalanobis distance is calculated to reject outliers that exceed a
probabilistic threshold
\begin{equation}
    \chi^2 = \mathbf{y}^T \mathbf{S}^{-1} \mathbf{y}
\end{equation}
Measurements that pass the threshold test are used to update the state
\begin{equation}
    \hat{\mathbf{x}} = \hat{\mathbf{x}} \oplus \mathbf{K} \mathbf{y}
\end{equation}
Where the state composition $\oplus$ is defined as addition for vectors and
quaternion multiplication for quaternions. Additionally, the covariance update
is computed with the following
\begin{equation}
    \mathbf{P} = (\mathbf{I} - \mathbf{K}\mathbf{H}) \mathbf{P}(\mathbf{I} - \mathbf{K}\mathbf{H})^T + \mathbf{K}\mathbf{R}\mathbf{K}^T
\end{equation}
Known as the Joseph form, it is used for it's balance of improved numerical
stability with computational complexity \cite{Yaakov}.

\section{Experimental Results}

\subsection{One-Way Time Filtering}

\begin{figure}
    \centering
    \includegraphics[width = 0.9\linewidth]{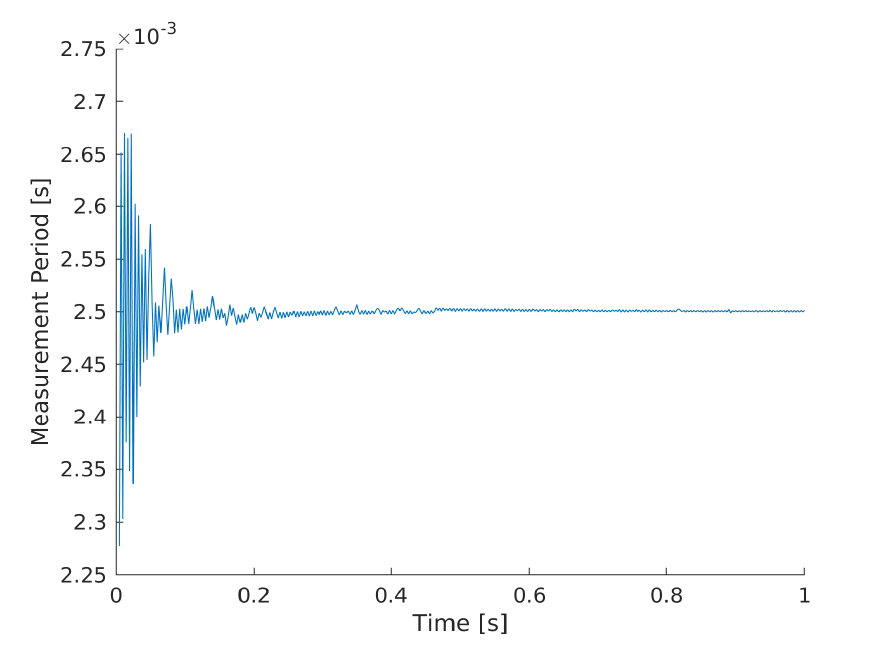}
    \caption{Measurement period translated from sensor time shown converging to 400 Hz ($2.5 \times 10^{-3}$ seconds)}
    \label{fig:delta}
\end{figure}

For both the VectorNav IMU and the Basler Ace camera, we implemented the one-way
time translation filter in order to improve the inconsistent time measurements.
The two-parameter filter converges quickly for both parameters.
This led to rapid reduction in noise reported in the
sensor measurements, with convergence to the expected measurement frequency, as
shown in \cref{fig:delta}. Similar convergence of the parameters and the
measurement time step for the USB camera were observed. Specifically, the translated
measurement period settled within 10 $\mu s$ of the reported sensor measurement period
in 0.3 seconds for the IMU and 1.0 seconds for the USB camera. Compared to
the convergence time on the order of seconds or even minutes seen in PTP
filters, we believe this is a reasonable translation convergence time to see.

Due to the convergence in the differences of the sensor time output in each the
USB IMU and camera, we find that the application of this one-way time filter to
be useful in rejecting the noise that can appear when using non-networked
sensors. Additionally, due to the low overhead of the calculations, even high
rate sensors, such as IMUs, can use this filter without substantial reductions
in throughput.

\subsection{Multi-Camera EKF Results with Visual Overlap}

\begin{figure}
    \centering
    \includegraphics[width = 0.9\linewidth]{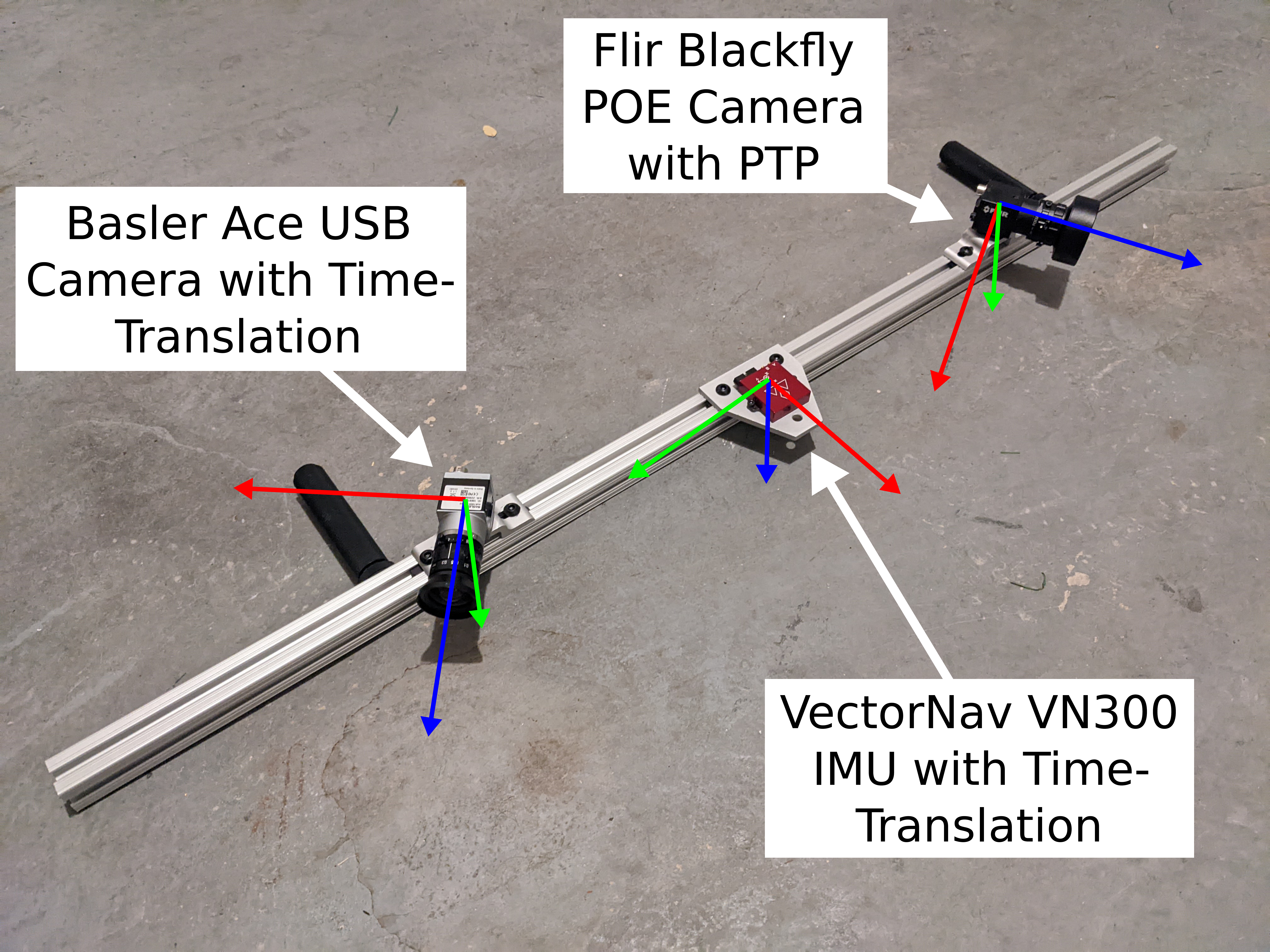}
    \caption{
        Mounting system used to test multi-camera IMU calibration with
        configuration showing non-overlapping fields of view
    }
    \label{fig:CalibrationRig}
\end{figure}

To validate the efficacy of the filter design, we used a multi-camera IMU rig
that is capable of rigidly mounting cameras and an IMU in various setups shown
in \cref{fig:CalibrationRig}.
Using this rig, it was possible to test the filter using both PoE and USB 3.0
cameras and differing rates.

Using a stationary ChArUco calibration target, data was collected exciting all
six degrees of freedom
while keeping the calibration pattern in the field of view for both cameras.
The time translation filter operated on the USB sensors, and the calibration
filter was applied on the sensor output and camera detections using the OpenCV
estimatePoseBoard function. We found that the filter consistently converged on
sensor extrinsic parameters given the measurements and typically within 20
seconds of the initial detections of the calibration target.

\begin{figure}
    \centering
    \includegraphics[width = 0.9\linewidth]{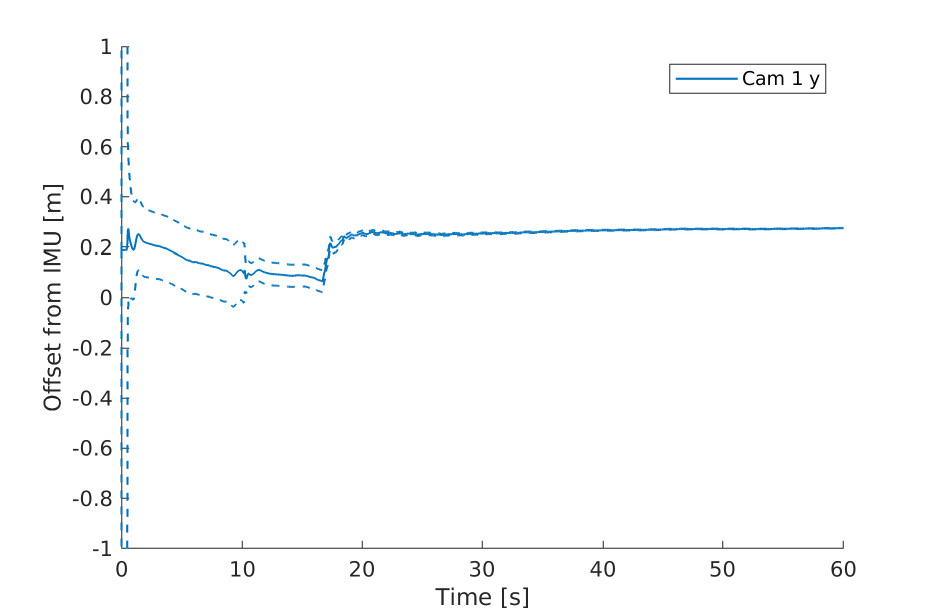}
    \caption{Camera 1 independent y-axis convergence plot}
    \label{fig:Cam1Y}
\end{figure}

\begin{figure}
    \centering
    \includegraphics[width = 0.9\linewidth]{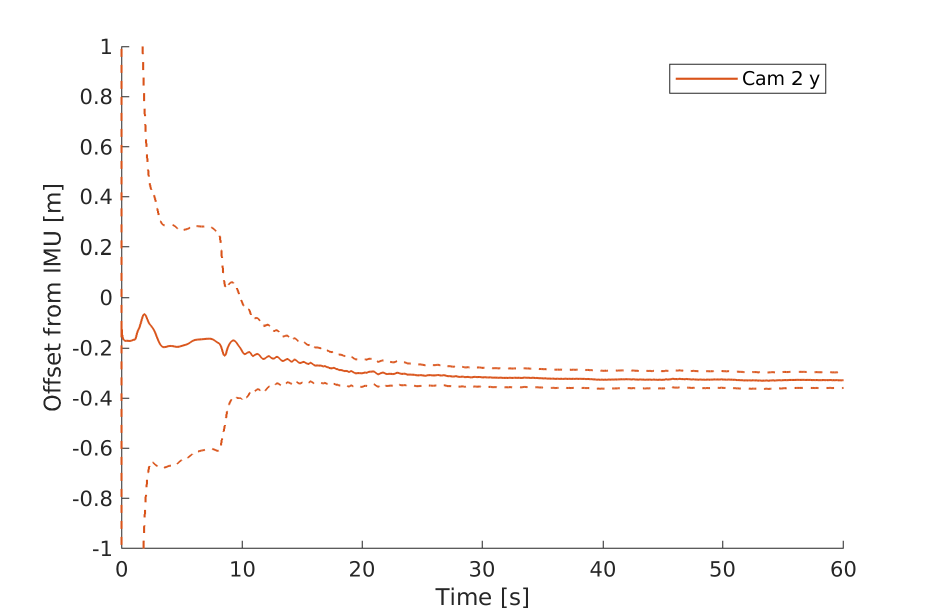}
    \caption{Camera 2 independent y-axis convergence plot}
    \label{fig:Cam2Y}
\end{figure}

\begin{figure}
    \centering
    \includegraphics[width = 0.9\linewidth]{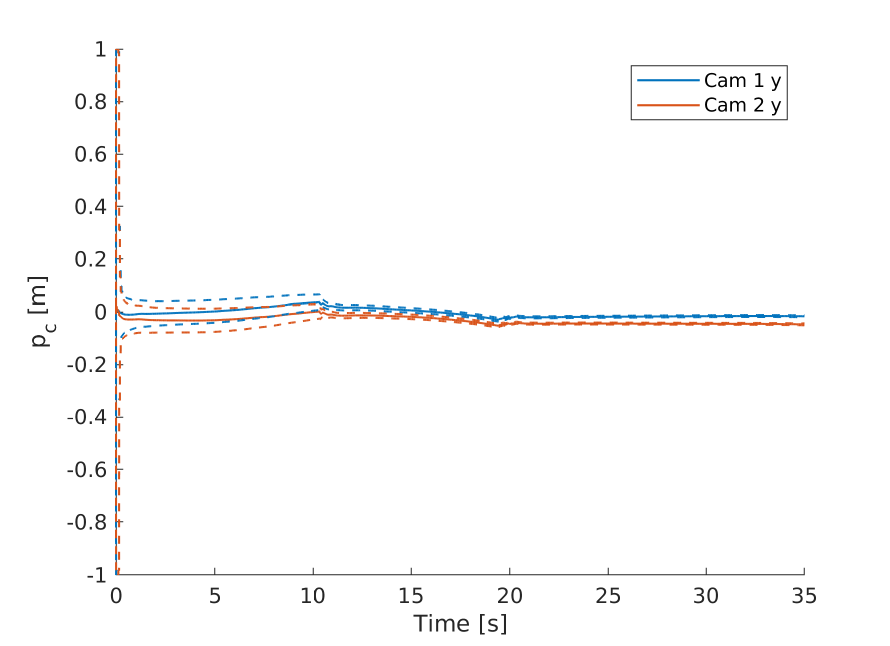}
    \caption{Dual camera y-axis convergence plot with overlapping fields of view}
    \label{fig:DualCamY}
\end{figure}

Additionally, it was possible to run the filter with either one of the camera
measurement streams. Shown is the convergence of the main axis of offset between
the cameras in the IMU frame. In this case, that is the y-axis.
\cref{fig:Cam1Y,fig:Cam2Y} show the convergence of the
camera extrinsics when only a single camera is used in the filter. When both
cameras are used, shown in \cref{fig:DualCamY}, the
filter converges faster and more consistently. Similar results were seen in the
remaining translation and rotation axes.

\subsection{Filter Evaluation}
In order to validate the implementation of our algorithm, we compare our
experimental results of with the outputs the MatLab camera calibration
toolbox \cite{MatLabCamCalib}. Using synchronized captures from both cameras, a
relative offset was found that could be compared to the output of the filter,
shown in \cref{tab:StereoResults}.
The relative difference between the calibration application and the EKF is quite
low, especially with regards to the physical location. We note the Y offset
(The direction into the camera frame) had higher error than the other
dimensions, which we believe is attributable to the small calibration target
used that limited the excitation possible in this direction. With this in mind,
we believe the converged filter values to be validated.

\begin{table}
    \centering
    \caption
    {
        Comparison of extrinsic parameters in IMU frame
        from MatLab Stereo Calibrator and
        the multi-camera IMU EKF with overlapping fields of view
    }
    \label{tab:StereoResults}
    \begin{tabular}{c|c|c|c}
        Dimension & Units & MatLab & EKF   \\
        \hline
        X         & mm    & 500    & 497   \\
        Y         & mm    & -11.8  & -30.8 \\
        Z         & mm    & -53.1  & -56.9 \\
        $\alpha$  & deg   & 0.80   & 2.06  \\
        $\beta $  & deg   & -6.29  & -9.00 \\
        $\gamma$  & deg   & -3.21  & -5.33
    \end{tabular}
\end{table}

\subsection{Multi-Camera EKF Results without Visual Overlap}

\begin{figure}
    \centering
    \includegraphics[width = 0.9\linewidth]{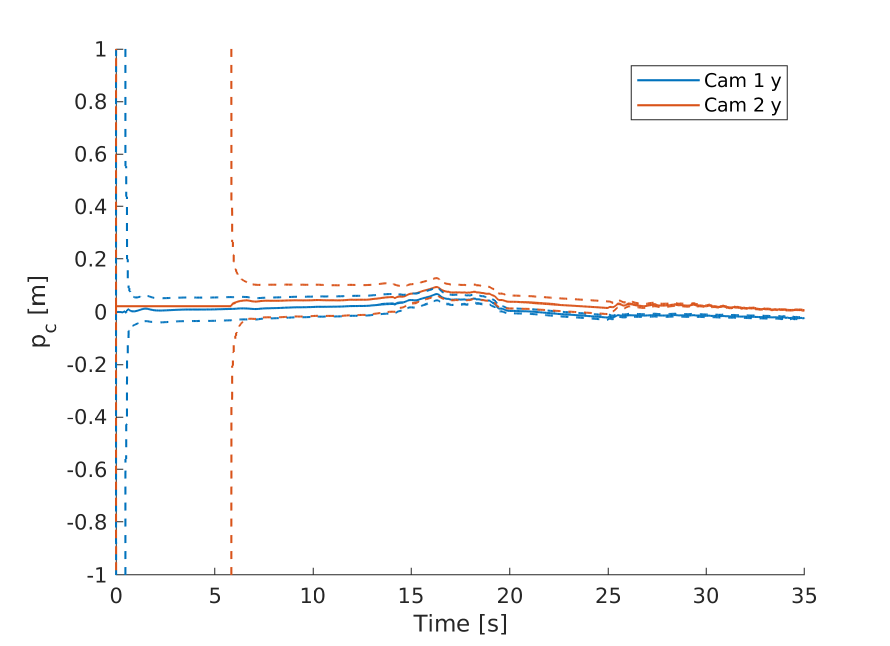}
    \caption{Dual camera y-axis convergence plot with non-overlapping fields of view}
    \label{fig:DualCam_Y_NoOverlap}
\end{figure}

A nice feature of this structure of Kalman filter for calibrating multiple
cameras is that there is no presumption of overlapping fields of view or
visibility of targets throughout the calibration procedure. In fact, we show
experimentally that cameras facing opposing directions still lead to
convergent calibrations to nominally correct values when at some point
throughout the calibration procedure the cameras are shown the calibration
target. A selection of the results of such an experiment are shown in
\cref{fig:DualCam_Y_NoOverlap}
where the y-axis shows convergence and each camera converges separately
when the target is within its own field of view. Moreover, because the separate
cameras are rigidly linked and are each correlated with the IMU, we see the
interesting result that a camera not currently looking at the calibration target
may still experience a reduction in uncertainty due to the measurements of a
separate camera.

\section{Conclusions}

In this paper, we have presented an extension to an EKF-based extrinsic
parameter calibration for multiple camera and IMU systems that can be quickly
implemented using a variety of existing software for fiducial detection
as well as address the issue of time translation when
closed loop or hardware synchronization is not possible.

The time translation estimator was tested for both IMU and camera
sensors, and showed convergence in both cases. Fast convergence was seen from
the sensors using the time translators, which overall reduced the timing noise.

The estimator was tested experimentally with single camera-IMU and dual
camera-IMU setups with overlapping fields of view. The estimator converged for
all extrinsic calibration parameters in both cases, and the values of one set of
parameters was confirmed using a stereo camera calibration tool.

Additionally, experiments were made with non-overlapping fields of view to show
the estimator is also capable of converging in these scenarios and is robust
against periods without measurement data. The EKF framework showed additional
advantages in this type of scenario as cameras not producing measurements still
had a reduction in covariance due to the detections of other cameras.


%





\ifCLASSOPTIONcaptionsoff
    \newpage
\fi



\bibliographystyle{IEEEtran}
\bibliography{references}

\end{document}